\documentclass[review,10pt,nopreprintline]{elsarticle} 
\usepackage{float}
\usepackage{amsmath}
\usepackage{graphicx}
\usepackage{amsfonts}
\usepackage[unicode=true, bookmarks=false]{hyperref}
\usepackage[latin9]{inputenc}
\usepackage{times} 
\usepackage[verbose,tmargin=1in,bmargin=1in,lmargin=1in,rmargin=1in]{geometry}
\usepackage{setspace} 
\usepackage[superscript]{cite}
\usepackage{xcolor}
\usepackage{titlesec}
\titleformat*{\section}{\large\bfseries\sffamily}
\titleformat*{\subsection}{\normalsize\bfseries\sffamily}
\titleformat*{\subsubsection}{\small\bfseries\sffamily}

\renewenvironment{abstract}{\global\setbox\absbox=\vbox\bgroup
  \hsize=\textwidth%
  \noindent\unskip\textbf{\large Summary}
  \par\medskip\noindent\unskip\ignorespaces}{\egroup}
\makeatletter
\renewcommand\@biblabel[1]{#1} 
\makeatother

\begin{document}
\begin{frontmatter}
\title{Towards Cellular-Scale Interpretability in Pathology Foundation Models for Biomarker Assessment}


%

\author[1,2,3,10]{Jingsong Liu\corref{corrauth}\fnref{equalcont}}
\ead{jingsong.liu@tum.de}

\author[2,3,4]{Han Li\fnref{equalcont}}

\author[1,2,3]{Zhengyang Xu}
\author[1]{Franz-Leonard Klaus}
\author[1]{Fabian St\"ogbauer}
\author[5]{Shihui Zu}
\author[5]{Weiwei Zhou}
\author[1]{Atsuko Kasajima}
\author[1]{Felix Schicktanz}
\author[1]{Alexander Muckenhuber}
\author[1]{Julius Shakhtour}
\author[6]{Jiale Yu}
\author[2]{Tiannan Zheng}
\author[2]{Xun Ma}
\author[2]{Maggie Wang}
\author[1,2,3,10]{Christian Grashei}
\author[8]{Bao Li}

\author[9]{Guiyang Jiang}
\author[5]{Hongming Xu}
\author[7]{Shaohua Kevin Zhou}
\author[2,3,4]{Nassir Navab}
\author[1,2,3,10]{Peter Sch\"uffler}

\address[1]{Institute of Pathology, Technical University of Munich, Munich, Germany}

\address[2]{School of Computation, Information and Technology, Technical University of Munich, Munich, Germany}

\address[3]{Munich Center for Machine Learning (MCML), Munich, Germany}

\address[4]{Computer Aided Medical Procedures (CAMP), Technical University of Munich, Munich, Germany}

\address[5]{School of Biomedical Engineering, Faculty of Medicine, Dalian University of Technology, Dalian, China}

\address[6]{Affiliated Hospital of Chifeng University, Chifeng, China}

\address[7]{Center for Medical Imaging, Robotics, and Analytic Computing \& Learning (MIRACLE), Suzhou Institute for Advanced Research, USTC, Suzhou, China}

\address[8]{Department of Biomedical Informatics, Harvard Medical School, Boston, MA, USA}

\address[9]{The First Hospital and the College of Basic Medical Sciences of China Medical University, Shenyang, China}

\address[10]{Munich Data Science Institute (MDSI), Munich, Germany}

\fntext[equalcont]{These authors contributed equally to this work.}
\cortext[corrauth]{Corresponding author}

\begin{abstract}

Molecular biomarker testing in pathology is often costly and tissue-consuming, limiting scalable clinical deployment. Artificial intelligence applied to hematoxylin and eosin (HE)-stained histology could enable rapid biomarker screening, but clinical translation requires models that are both accurate and interpretable. Here we introduce Hireca, a biomarker-focused pathology foundation model pretrained on more than 80,000 whole-slide images spanning 38 organ types from three medical centers, together with CytoMap, an interpretability module that localizes cellular-scale evidence underlying predictions. Across 10 biomarker tasks encompassing morphological, molecular, genetic, and spatial-transcriptomic-proxy readouts, Hireca ranked first in five tasks and outperformed comparable models overall. In evaluation by eight pathologists from two countries, CytoMap was consistently preferred over alternative visualization approaches and revealed error patterns in difficult cases. These results position Hireca and CytoMap as a transparent framework for clinically reviewable biomarker assessment directly from routine HE histology.


\end{abstract}


\end{frontmatter}

\section*{Introduction}

Biomarkers are measurable molecular, protein, or cellular characteristics that reflect biological processes, disease states, or responses to therapy \cite{cagney2018fda}. Across oncology, they play a central role in disease classification, risk stratification, therapeutic selection, and treatment monitoring \cite{aldoughaim2024cancer}. While some
histomorphological biomarkers, such as mitotic figures, can be directly assessed from haematoxylin and eosin (HE) histopathology images, many clinically relevant biomarkers typically require assay-specific methods, including immunohistochemistry (IHC), polymerase chain reaction (PCR), in situ hybridization (ISH), and sequencing-based tests~\cite{lindeman2018updated,mosele2020recommendations}.

Breast cancer provides a representative example of biomarker assessment. Approximately four out of five breast cancer cases are hormone-receptor positive \cite{walsh2020management}, and estrogen receptor (ER), progesterone receptor (PR), human epidermal growth factor receptor 2 (HER2), and Ki-67 are routinely evaluated by IHC to guide treatment planning and therapeutic stratification \cite{allison2020estrogen}. Among these biomarkers, the clinical interpretation of HER2 is particularly dependent on cell-level features, including the completeness and intensity of membranous staining in invasive tumor cells and the proportion of cells exhibiting these patterns \cite{american2007breast,ivanova2024standardized}.


Recent computational pathology studies have shown that some clinically relevant biomarkers, including molecular \cite{farahmand2022her2,akbarnejad2025toward}, genetic \cite{kather2019histological}, and spatially resolved transcriptomic biomarker states \cite{jaume2024hest}, can also be inferred directly from routine HE images. However, this setting is fundamentally different from routine IHC or molecular testing. Rather than directly measuring the target biomarker, HE-based prediction relies on indirect morphological correlates that may be subtle and difficult to recognize during visual review. 
Compared to many other diagnostic and prognostic tasks, such as survival prediction or cancer subtyping, HE-based biomarker assessment is therefore a clinically important yet particularly demanding setting: it requires not only accurate prediction but also interpretable localization of the morphological evidence at a spatial scale aligned with pathological review. The relevant evidence may arise from sparse, localized cellular regions and be expressed through subtle cellular morphology or interactions between tumor cells and the surrounding microenvironment~\cite{li2022clinical}.

Recent advances in computational pathology have shifted the field from task-specific models toward pathology foundation models (PFMs), which serve as general-purpose backbones for histopathological image analysis. While task-specific approaches, including CNN-based models \cite{he2016deep}, multiple-instance learning methods \cite{farahmand2022her2}, and cell-centric models \cite{graham2019hover,li2025nuhtc}, have shown promise for specific diagnostic or biomarker-related tasks, they are often optimized for particular tissues, endpoints, or datasets. In contrast, representative PFMs \cite{chen2024towards,zimmermann2024virchow2,lu2024visual,vorontsov2024foundation}, pretrained on large-scale collections of HE tissue patches, have demonstrated stronger transferability and generalization across diverse downstream tasks, tumor types, and clinical cohorts.

However, despite their strong predictive performance, it remains unclear whether PFMs capture the fine-grained morphological features required for biomarker assessment. Most existing evaluations focus on slide-level classification accuracy or coarse regional prediction, without explicitly examining whether the learned representations encode cell-level phenotypes, tumor-cell morphology, or microenvironmental patterns. 


Beyond predictive performance, a further limitation concerns the spatial granularity of model interpretation. For CNN-based methods, commonly used attribution approaches such as GradCAM \cite{selvaraju2020grad} typically provide coarse regional localization, limiting their value for cell-level interpretation. For Transformer-based PFMs, attention maps are frequently used to visualize model focus, but they likewise do not readily provide high-resolution, pathologically interpretable evidence for biomarker assessment \cite{darcet2023vision,simeoni2025dinov3,shi2026vision}. This limitation is particularly critical for biomarker-related tasks, where diagnostically relevant signals may arise from a small number of decisive cells, subtle cellular morphology, or fine-grained microenvironmental interactions \cite{li2022clinical}. At a typical scanning resolution of 0.5~$\mu$m/pixel (20$\times$), a cell nucleus or a cellular region on the order of 5--20~$\mu$m corresponds to about 10--40 pixels \cite{lammerding2011mechanics}. As a result, existing interpretability strategies remain insufficient for assessing whether models capture biomarker-relevant morphology at near-cellular scale.

\begin{figure}[H]
    \centering
    \includegraphics[width=.9\textwidth]{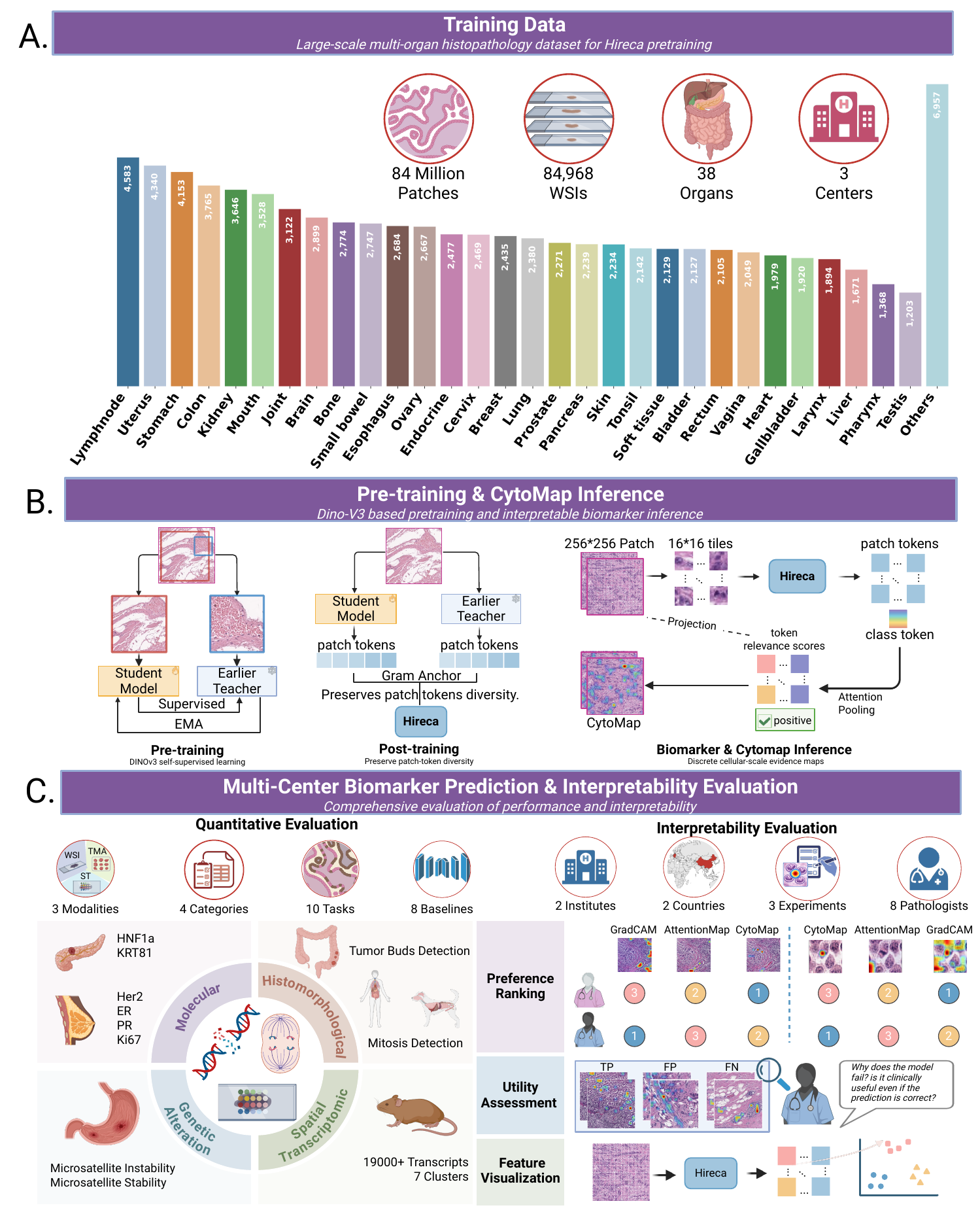} 
    \caption{\textbf{Overview of the Hireca's pretraining and evaluation pipeline.}
\textit{A) Training data:} Hireca was pretrained on 84,968 whole-slide images from 3 medical centers, covering 38 organs with broad multi-organ distribution.
\textit{B) Pretraining and CytoMap inference:} Hireca is pretrained using the DINOv3 self-supervised framework, with a special post-training step that preserves patch-token diversity. During biomarker inference, CytoMap maps patch-token relevance scores back to the input HE patch, producing discrete cellular-scale evidence maps for biomarker interpretation alongside biomarker prediction.
\textit{C) Multi-center biomarker prediction and interpretability evaluation:} Biomarker prediction was benchmarked across 10 tasks spanning 4 categories, including histomorphological,  molecular, genetic alteration, and spatial transcriptomic proxy tasks, across 3 data modalities and against 8 FMs. Cellular-scale interpretability was further evaluated with 8 pathologists across 2 countries using three complementary evaluation components: (1) preference-based assessment, (2) clinical-utility and failure-case assessment, and (3) pathologist-observable and representation-space analyses. Abbreviations: WSI, whole-slide image; TMA, tissue microarray; ST, spatial transcriptomics.}
\label{Figure:introduction}
\end{figure}

To address these challenges, we propose Hireca, a High-Resolution pathology foundation model with Cellular-scale interpretability. Technically, Hireca differs from existing pathology foundation models in the DINOv3 pretraining framework. This design enables Hireca to preserve diverse cellular-scale patch-token representations rather than relying only on coarse patch-level features. As a result, Hireca can better capture cellular and microenvironmental cues that are essential for biomarker prediction, while also providing high-resolution cellular-scale interpretability through its visualization module, CytoMap (Fig. \ref{Figure:introduction}). 

To comprehensively evaluate Hireca, we introduce a clinically grounded benchmark. This benchmark assesses two complementary axes: quantitative biomarker prediction and interpretability quality, combining both public datasets for reproducible comparison and internal real-world diagnostic cohorts to strengthen clinical relevance. It includes ten biomarker tasks across four categories, covering histomorphological, molecular, genetic alteration, and spatial transcriptomic proxy settings. 


In the quantitative biomarker prediction evaluation,  Hireca achieved the best-ranked results on 5 out of 10 tasks, the highest number among the nine compared foundation models.
For interpretability quality, CytoMap was assessed through pathologist preference-based and clinical score-based evaluations involving eight pathologists from Europe and Asia, including four senior and four junior readers. These evaluations showed consistent expert preference for CytoMap over alternative visualization methods and demonstrated its utility in supporting HER2 clinical interpretation. Beyond these evaluations, failure-case analysis showed that CytoMap can expose model error patterns and reveal the current limits of Hireca in challenging biomarker settings. Finally, pathologist-observable evidence and representation-space analyses further showed that CytoMap interpretations are supported by biologically meaningful cellular-scale evidence. Such fine-grained morphological patterns are aligned with pathological assessment and enable Hireca to move beyond conventional slide-level prediction and patch-level interpretability toward cellular-scale interpretability for biomarker assessment.

\section*{Results}

The biomarker-oriented pathology foundation model, Hireca, was built on the recent self-supervised DINOv3 architecture and trained on more than 80,000 multi-center whole-slide images spanning 38 organs. We evaluated Hireca through a stepwise evidence chain linking biomarker prediction performance and interpretability. For prediction performance, Hireca was benchmarked against eight state-of-the-art foundation models across ten biomarker tasks covering four categories: histomorphological assessment, molecular biomarkers, genetic alteration biomarkers and spatial transcriptomic proxy biomarkers. For interpretability, CytoMap was assessed through two pathologist-centered evaluations: a preference-based blinded ranking study against alternative visualization methods, and a utility-based HER2 assessment examining whether CytoMap supports both case interpretation and error understanding. We further used failure-case analysis to characterize model error patterns and boundary conditions. Finally, we evaluated whether CytoMap provides cellular-scale interpretability through pathologist-observable examples and representation-space analysis.

\subsection*{Hireca achieved the highest number of best-ranked results across an extensive benchmark}

\begin{figure}[htbp]
\centering
\includegraphics[width=\textwidth]{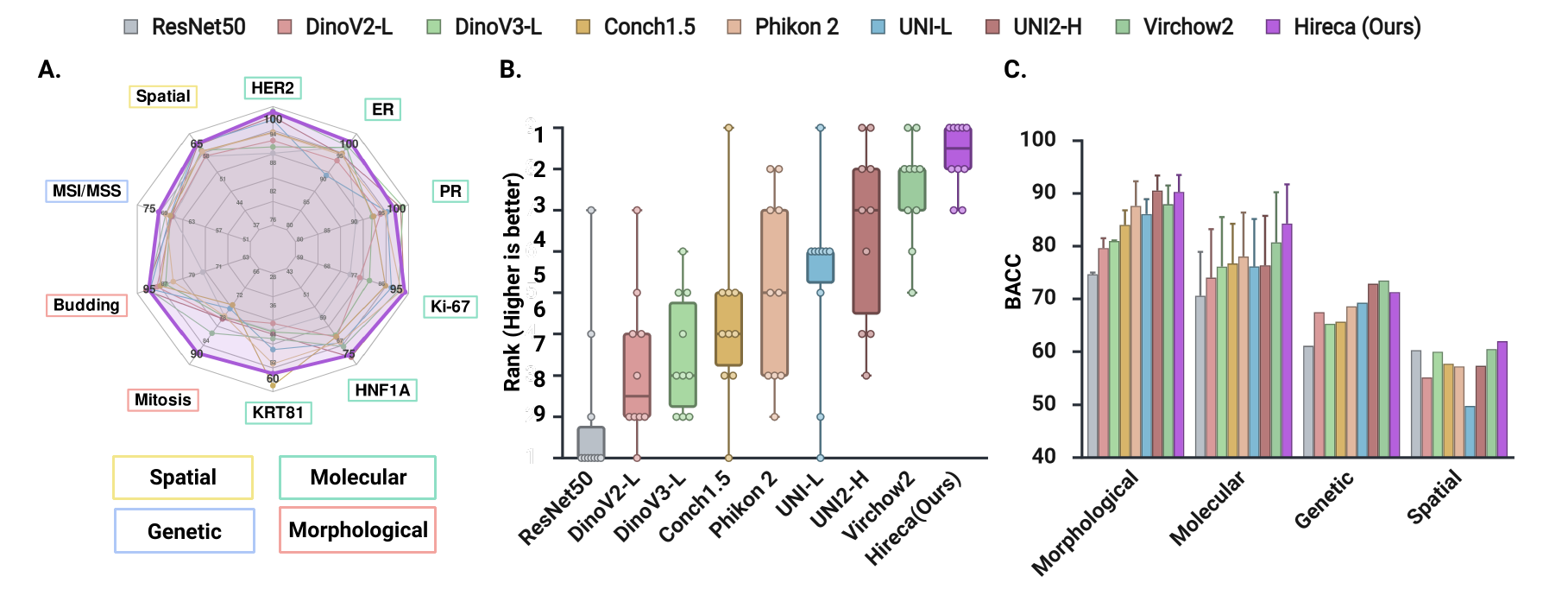} 
\caption{\textbf{Hireca vs Eight state-of-the-art methods across three experiments covering four biomarker categories.}
\textit{A) Radar graph of the task-level balanced accuracy (BACC) across ten biomarker tasks:} Hireca achieves the highest BACC on multiple tasks and remains competitive across the full benchmark.
\textit{B) Overall ranking across ten biomarker tasks:} The rank is based on BACC, a higher rank indicates better performance. Hireca attains the best aggregate ranking among all compared baselines.
\textit{C) Mean BACC across 4 biomarker categories:} Hireca shows the strongest average performance in the histomorphological, molecular, and spatial transcriptomic proxy categories, and remains competitive in the genetic alteration biomarker setting despite using a smaller backbone than some leading baselines.}\label{Figure:benchmark_result}
\end{figure}

\noindent{}\textbf{Histomorphological Assessment.}
Histomorphological assessment is important because it evaluates whether pathology foundation models can capture visually explicit and diagnostically meaningful histomorphological patterns. Hireca has been challenged on two morphology-driven tasks, mitosis detection and tumor budding.  Mitoses reflect tumor proliferative activity, whereas tumor budding captures invasive growth at the tumor front; both are medically relevant patterns that require recognition of localized cellular or small glandular structures in HE images. Hireca ranked first in tumor budding detection (balanced accuracy [BACC] = 0.935) and second in mitosis detection (BACC = 0.871), slightly below UNI2-H (BACC = 0.877), despite using a backbone approximately half its size. Overall, Hireca achieved the best ranking across the two histomorphological tasks (Fig. \ref{Figure:benchmark_result}).

\noindent\textbf{Molecular Biomarker Detection.}
We next assessed molecular biomarker detection to determine whether Hireca could detect therapeutically and biologically relevant molecular biomarker states from HE images. Hireca was evaluated on six molecular biomarkers across two distinct settings: four breast cancer biomarkers predicted from HE patches (HER2, ER, PR, and Ki67) and two PDAC biomarkers evaluated on tissue microarray (TMA) sections (HNF1a and KRT81; Sec. \ref{molecular_dataset}). 
Across these six molecular biomarker tasks, Hireca achieved the highest number of top-ranked results, ranking first on three tasks (3/6) (Fig.~\ref{Figure:benchmark_result} B). In the breast cancer HE setting, Hireca ranked first on ER (BACC = 0.982) and PR (BACC = 0.952), achieved a near-best result on HER2 (BACC = 0.987, compared with 0.989 for Virchow2), and remained competitive on Ki67 (BACC = 0.904). In the PDAC TMA setting, Hireca ranked first on HNF1a (BACC = 0.703) and second on KRT81 (BACC = 0.539). Together, these results show that Hireca performs consistently across different molecular biomarkers and tissue formats.


\noindent\textbf{Genetic Alteration Detection.}
Genetic alteration biomarker detection represents a  challenging setting because the target signals are often indirect and difficult to assess by visual inspection. For MSI/MSS classification in gastric cancer (Sec. \ref{msi_dataset}), Hireca ranked first among models with a comparable backbone size (including UNI-L, CONCH 1.5, and Phikon-v2). Overall, Hireca ranked third (BACC = 0.713), following the larger-backbone models UNI2-H (BACC = 0.729) and Virchow2 (BACC = 0.735) (Fig.~\ref{Figure:benchmark_result} A, C). These results suggest that Hireca captures genotype-associated histopathological cues with a more compact model design.

\noindent\textbf{Spatial Transcriptomic State Prediction.}
Finally, we evaluated a more demanding task: inferring spatially resolved gene-expression states from local tissue morphology in HE images. Using our in-house 10x Genomics Visium dataset of mouse pancreas with acinar cell-derived preneoplastic lesions (Sec.~\ref{st_dataset}), we assessed whether Hireca can distinguish local tissue niches defined by spatial transcriptomic profiles that are not directly visible in histological images. In this setting, Hireca achieved the highest BACC (0.620) among all models (Fig. \ref{Figure:benchmark_result} A, C), supporting its ability to capture subtle morphology-transcriptome associations and further probing the limits of information that can be inferred from routine HE morphology.

Overall, across an extensive and diverse biomarker prediction benchmark with tasks of varying complexity, Hireca achieved the highest number of top-ranked results, ranking first on 5 of 10 tasks, even when compared with substantially larger models. These findings suggest that Hireca's design, which preserves diverse cellular-scale cues, contributes to its strong performance across biomarker prediction tasks.

\begin{figure}[htbp]
\centering
\includegraphics[width=\textwidth]{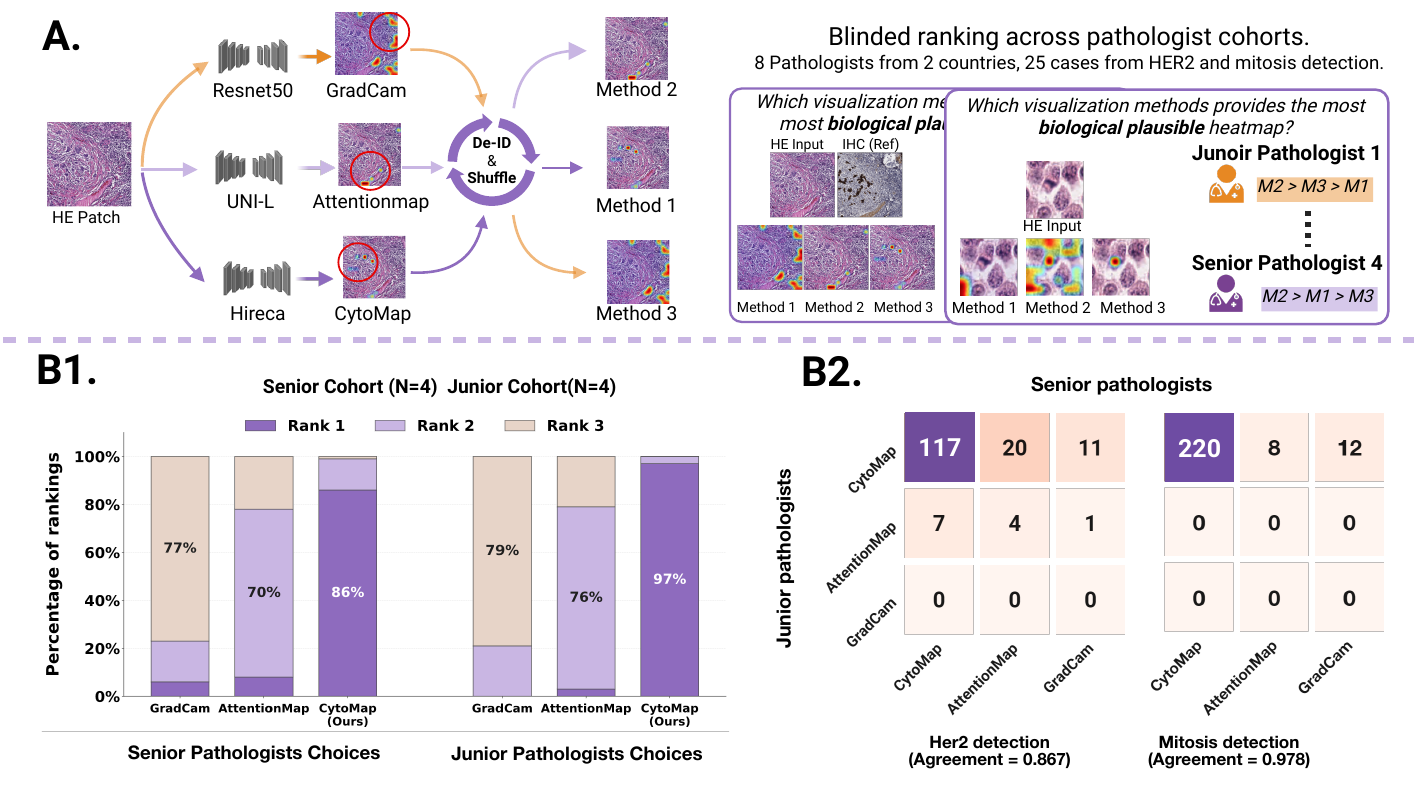} 
\caption{\textbf{ Pathologist preference-based interpretability evaluation of Hireca (CytoMap) against two state-of-the-art methods.} 
\textit{A) Multi-cohort blinded pathologist ranking workflow:} Visualization maps generated by GradCAM, AttentionMap, and CytoMap were de-identified and randomly shuffled as Method 1--3. Eight pathologists from Germany and China, including four senior and four junior pathologists, independently ranked the three methods across 25 cases from HER2 and mitosis detection according to biological relevance.
\textit{B1) Preference ranking results from junior and senior pathologist cohorts:} Stacked bar charts show the percentage distribution of Rank 1, Rank 2, and Rank 3 assignments in each cohort. CytoMap received the highest proportion of Rank 1 selections in both cohorts (86\% senior and 97\% junior), indicating consistent preference over the two state-of-the-art interpretability methods. 
\textit{B2) Agreement analysis between senior and junior pathologist cohorts:} Confusion matrices compare cohort-level rankings for HER2 and mitosis detection, showing strong diagonal concentrations and confirming that the preference pattern observed in B1 is reproducible across experience levels. The agreement scores were 0.867 for HER2 and 0.978 for mitosis detection.} \label{Figure:Ranking_experiment}
\end{figure}

\subsection*{Interpretable CytoMap was consistently preferred across pathologist cohorts.}

Despite Hireca's strong performances in biomarker prediction tasks, it remains unclear whether
Hireca truly capture fine-grained, diagnostically relevant morphological features required for clinical assessment.
 To assess the interpretability of CytoMap, Hireca's attention-based visualization module, we conducted a multi-cohort blinded pathologist ranking study against two state-of-the-art interpretability methods, GradCAM and AttentionMap. Visualization maps generated by the three methods were de-identified and randomly shuffled as Method 1--3. Eight pathologists from two countries, including four senior and four junior pathologists, independently ranked the methods across 25 cases from HER2 and mitosis detection according to biological relevance (Fig.~\ref{Figure:Ranking_experiment} A).

\noindent\textbf{CytoMap achieves the highest pathologist preference.}
Across the 25 HER2 and mitosis cases, CytoMap received the highest proportion of Rank~1 selections in both senior and junior cohorts (Fig.~\ref{Figure:Ranking_experiment} B1). Senior pathologists assigned CytoMap Rank~1 in 86\% of evaluations, while junior pathologists assigned CytoMap Rank~1 in 97\% of evaluations. In contrast, AttentionMap was most frequently ranked second, whereas GradCAM was most frequently ranked third in both cohorts. These results indicate that pathologists consistently preferred the cellular-scale evidence provided by CytoMap over the two state-of-the-art interpretability methods.

\noindent\textbf{CytoMap preference was highly consistent across experience levels.}
We further assessed whether the preference pattern was consistent between senior and junior pathologist cohorts (Fig.~\ref{Figure:Ranking_experiment} B2). For both HER2 and mitosis detection, the cohort-level rankings showed strong diagonal concentrations, indicating that the preference for CytoMap was reproducible across different levels of pathology experience. This consistency was supported by high agreement scores of 0.867 for HER2 and 0.978 for mitosis detection. Together, these findings identify CytoMap as an expert-preferred interpretability method for cellular-scale biomarker evidence.




\subsection*{CytoMap demonstrates comprehensive clinical utility and interpretable failure boundaries through pathologist-in-the-loop assessment.}

\begin{figure}[htbp]
\centering
\includegraphics[width=.9\textwidth]{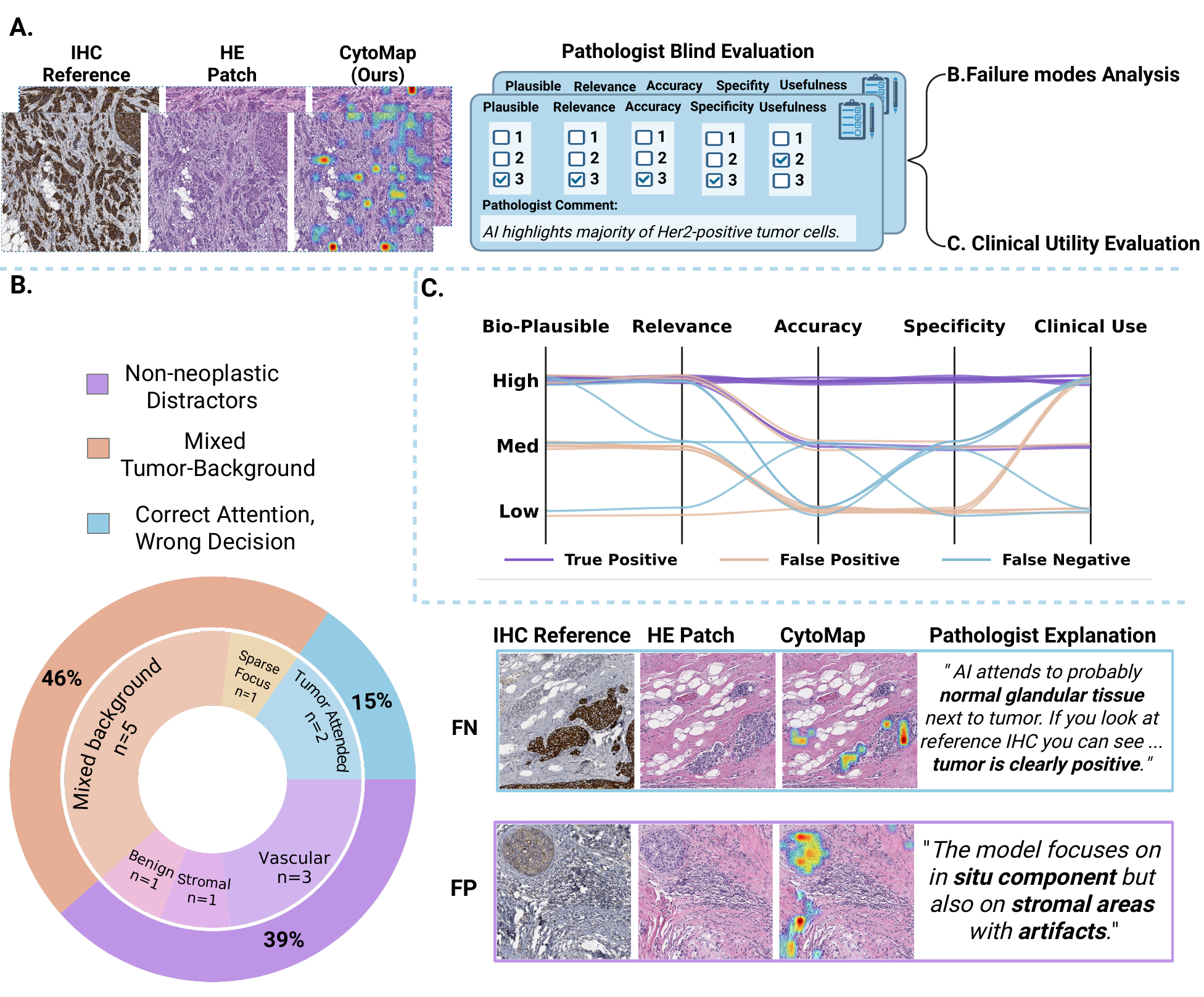} 
\caption{\textbf{Pathologist score-based interpretability evaluation of Hireca (CytoMap) for HER2 assessment.} \textit{A) Pathologist-in-the-loop assessment workflow:} For HER2 true-positive (TP), false-positive (FP), and false-negative (FN) cases, pathologists reviewed the IHC reference, the corresponding HE patch, and the CytoMap visualization. They scored the clinical utility of the highlighted evidence across multiple dimensions and provided comments explaining the reasons for successful or failed predictions. These scores and comments were further used for clinical utility evaluation and failure-case analysis.
\textit{B) Failure-case analysis:} FP and FN cases were reviewed by pathologists and categorized according to their dominant failure reason, including non-neoplastic distractors, mixed tumor-background attention, and correct attention with incorrect model decision. Colours denote different failure categories, and two representative examples are shown on the right.
\textit{C) Multidimensional clinical utility evaluation:} All cases were scored by pathologists using the prediction results and IHC reference across five dimensions: biological plausibility, relevance, accuracy, specificity, and clinical usefulness. Notably, clinical usefulness remained relatively high across case types, suggesting that cellular-scale CytoMap visualizations may support both correct case interpretation and error understanding.}
\label{Figure:Failure_analysis}
\end{figure}

Following the preference-based ranking study, we further examined whether CytoMap satisfies the practical utility requirements of providing useful evidence for correct predictions and supporting error understanding in incorrect predictions. We therefore performed a complementary pathologist-in-the-loop evaluation for HER2 assessment, including true-positive, false-positive, and false-negative cases. For each case, pathologists reviewed the IHC reference, the corresponding HE patch, and the CytoMap visualization. They scored the clinical utility of the highlighted evidence across five dimensions, including biological plausibility, relevance, accuracy, specificity, and clinical usefulness, and provided comments explaining the reasons for successful or failed predictions (Fig.~\ref{Figure:Failure_analysis}A).

\noindent{}\textbf{Failure-mode analysis exposes boundary conditions for Hireca.}
We first analyzed FP and FN cases to determine whether CytoMap could help explain model errors (Fig.~\ref{Figure:Failure_analysis} B). Pathologist comments showed that these errors followed structured and interpretable patterns rather than arbitrary activations. FP and FN cases were categorized according to their dominant failure reason, including mixed tumor-background attention, non-neoplastic distractors, and correct attention with incorrect model decision. These patterns accounted for 46\%, 39\%, and 15\% of reviewed error cases, respectively. In representative FP cases, CytoMap often highlighted tumor-associated regions but also included stromal, artefactual, or non-neoplastic components, indicating limited specificity and contextual confusion. In representative FN cases, CytoMap sometimes focused on insufficiently informative regions despite HER2-positive tumor cells being visible in the IHC reference. These results show that CytoMap can expose model error patterns and reveal the current boundaries of Hireca in challenging biomarker assessment. Such interpretable failure analysis is essential for understanding where HE-based biomarker models can be clinically trusted and where additional validation remains necessary.

\noindent{}\textbf{CytoMap supports case interpretation and error understanding.}
We next evaluated whether the highlighted evidence by CytoMap was clinically useful for case-level interpretation across both correct and wrong cases (Fig.~\ref{Figure:Failure_analysis}B2). In TP cases, CytoMap highlighted compact tumor-cell regions that were scored highly across biological plausibility, relevance, accuracy, specificity, and clinical usefulness. These findings indicate that, although HER2 status is conventionally established by IHC, CytoMap can localize pathologically meaningful HE-based evidence associated with HER2-positive tumor regions. Importantly, clinical usefulness remained relatively high across case types, suggesting that CytoMap can support both interpretation of correct predictions and also error understanding in incorrect predictions by making the model's evidence visible to pathologists.
Together, these findings support CytoMap not only as an explanation tool for successful predictions, but also as a practical interface for identifying failure modes before clinical translation.

\subsection*{CytoMap provides pathologist-observable and representation-space cellular-scale evidence.} 

\begin{figure}[htbp]
\centering
\includegraphics[width=.9\textwidth]{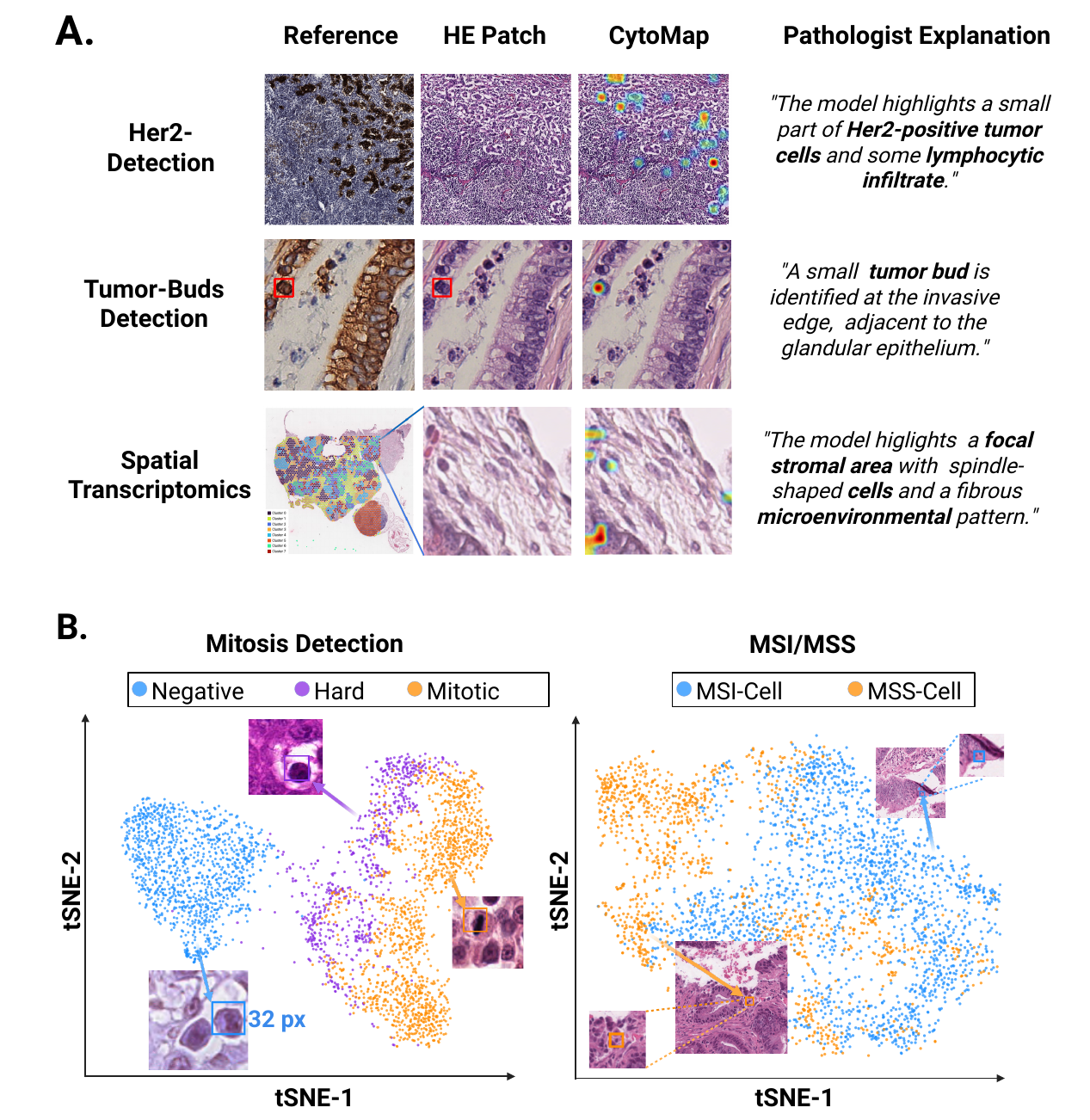} 
\caption{\textbf{CytoMap provides pathologist-observable and representation-space cellular-scale evidence.}
\textit{A) Pathologist-observable evidence for cellular-scale interpretability:} Representative examples across HER2 detection, tumor-bud detection, and spatial transcriptomic proxy prediction show that CytoMap highlights fine-grained cellular or microenvironmental structures that can be directly interpreted by pathologists.
\textit{B) Representation-space evidence for cellular-scale interpretability:} T-SNE visualization of CytoMap-highlighted patch-token features shows task-relevant organization for mitosis detection and MSI/MSS prediction, suggesting that the highlighted regions correspond to meaningful cellular morphology rather than arbitrary visual attention.
}
\label{Figure:tsne}
\end{figure}

Having established expert preference and practical utility through two pathologist-centered evaluations, we next examined whether CytoMap's interpretability is supported by cellular-scale evidence. We assessed this from two complementary perspectives: whether highlighted regions were observable by pathologists as cellular or microenvironmental structures, and whether they showed task-relevant organization in representation space.

\noindent\textbf{CytoMap provides pathologist-observable cellular-scale evidence.}
Across representative biomarker settings, including HER2 detection, tumor-bud detection, and spatial transcriptomic proxy prediction (\ref{Figure:tsne} A), CytoMap highlighted compact local regions rather than coarse tissue areas. These regions corresponded to fine-grained cellular or microenvironmental patterns that could be directly interpreted by pathologists, supporting the pathologist-observable nature of CytoMap evidence at cellular-scale.

\noindent\textbf{CytoMap-highlighted tokens cluster by cellular phenotype in representation space.}
We next examined whether CytoMap-highlighted regions also preserve task-relevant structure in feature space. To this end, we extracted highly attended patch-token features and visualized their Hireca embeddings using t-SNE (Sec.~\ref{interpretability_setup}; Fig.~\ref{Figure:tsne} B). For mitosis detection, mitotic, negative, and hard-to-tell tokens showed clear phenotype-associated organization. Mitotic and negative tokens occupied distinct regions, whereas hard-to-tell tokens were distributed between them, consistent with a morphological continuum from non-mitotic to mitotic cellular appearances. Representative token crops further confirmed that these regions corresponded to meaningful cellular morphologies rather than arbitrary image patches.

For MSI/MSS prediction, CytoMap-highlighted tokens showed weaker but still observable organization between MSI-associated and MSS-associated regions. This more partial separation is consistent with the greater morphological heterogeneity of MSI/MSS prediction, where relevant evidence is likely distributed across local epithelial, stromal, and microenvironmental contexts rather than confined to a single visually explicit cellular phenotype. Together, these findings show that CytoMap-highlighted regions are not only pathologist-observable in image space, but also organized in representation space according to task-relevant cellular and microenvironmental patterns, supporting CytoMap as a cellular-scale interpretability module for biomarker-oriented Hireca.

\section*{Discussion}
\subsection*{Summary of findings}

To comprehensively evaluate Hireca, we introduce an extensive and clinically grounded benchmark covering ten biomarker tasks across four categories using both public datasets and internal real-world diagnostic cohorts. Quantitatively, Hireca achieved the highest number of top-ranked results (on 5 of 10 tasks) even when compared with substantially larger models, highlighting the value of preserving diverse cellular-scale cues for biomarker prediction. Qualitatively, pathologist preference-based and clinical score-based evaluations showed that CytoMap was consistently preferred across pathologist cohorts, demonstrated comprehensive clinical utility, and revealed interpretable failure boundaries. Finally, the cellular-scale interpretability of CytoMap was further supported by pathologist-observable and representation-space evidence.

\subsection*{Clinical relevance and translational potential}

Our results highlight three clinically relevant points. 
(1) \textbf{HE-based biomarker prediction should not be treated as a conventional slide-level classification problem.} HE-based biomarker prediction differs from conventional slide-level classification because its relevant evidence is often sparse, local, and linked to cellular, molecular, or microenvironmental changes. Therefore, a clinically useful biomarker model should provide not only a prediction, but also localized evidence that pathologists can examine. (2) \textbf{Cellular-scale interpretability is essential for making AI-based biomarker prediction understandable to pathologists.} Cellular-scale interpretability is important because coarse heatmaps may indicate suspicious regions, but often do not explain which cells or local tissue structures support the decision. By highlighting compact cellular-scale regions, CytoMap brings model evidence closer to the spatial scale of routine pathological review, helping pathologists assess whether the model is correct for histologically meaningful reasons. (3) \textbf{Cellular-scale reviewability is important for understanding model errors and limitations.} Cellular-scale reviewability allows pathologists to inspect whether model errors arise from biologically meaningful but misleading evidence or from irrelevant regions. In the HER2 evaluation, CytoMap helped reveal whether incorrect predictions were related to non-neoplastic tissue, mixed tumor-background attention, or insufficiently informative regions.

Together, these findings show that Hireca and CytoMap support a more transparent form of biomarker AI, where predictions can be inspected, interpreted, and clinically contextualized by pathologists.

%



\subsection*{Broadening the scope: beyond biomarker prediction}
Although this work focuses on biomarker-oriented pathology tasks, the utility of Hireca may extend beyond biomarker prediction. In supplementary experiments (Sec.1) on more general pathology benchmarks, including tissue classification, tumor detection, and disease severity grading, Hireca maintained top-tier performance without task-specific architectural modification. This suggests that Hireca learns a broadly transferable histopathological feature space rather than a narrow biomarker-specific representation. In this sense, biomarker prediction serves not as an isolated application, but as a demanding test of whether a foundation model can preserve diagnostically meaningful local information while remaining broadly useful across pathology tasks.




\subsection*{Limitations and future directions}
While histomorphological and molecular biomarkers have been performed on multi-institutional datasets, the genetic and spatial transcriptomic proxy tasks were evaluated only in single-cohort settings, and broader multi-center validation would be beneficial to establish robustness across wider biological variation.
We further aim to extend the current framework to larger-context oncology tasks, such as tumor grading, prognosis prediction, and survival modeling. Extending cellular-scale interpretability toward slide-level and case-level assessment will require more efficient inference while preserving fine-grained interpretability. In addition, slide-level domain shift across institutions \cite{liu2025hasd}, staining protocols, and scanning devices remains a persistent challenge for clinical translation.

\section*{Methods}\label{sec2}

\subsection*{Hireca Pretraining} \label{subsec:pretraining}

\subsubsection*{Data Preparation.}
The training data comprise approximately 84 million patches extracted from 84,968 in-house HE-stained whole-slide images (WSIs) scanned at 40 $\times$ magnification covering 38 organ types. Each WSI was first segmented using Otsu's thresholding \cite{otsu1979threshold} to isolate tissue regions. Non-overlapping patches of size (512 $\times$ 512) pixels were then extracted from the tissue regions at 40 $\times$ magnification and resized to (256 $\times$ 256) pixels, corresponding to an effective magnification of approximately 20 $\times$.

\subsubsection*{Staining Augmentation.}
A major challenge in computational pathology is domain shift, which arises from variations across pathology centers, including differences in staining protocols, scanner settings, and digitization procedures \cite{liu2025hasd}. To develop a general-purpose biomarker prediction model, the foundation model (FM) must be robust to such staining shifts to ensure reliable deployment across different hospitals and laboratories. For this reason, we apply random staining augmentation \cite{shen2022randstainna} to image patches during self-supervised pretraining. Specifically, random perturbations are sampled from a Gaussian distribution and applied to the LAB and HSV color channels of the patches.

\subsubsection*{Large-Scale Pretraining.}
Following \cite{simeoni2025cijo}, the large-scale training process consists of two stages: self-supervised pretraining and Gram-anchored post-training.  

\noindent\textbf{Pretraining stage.} 
The model is trained using a combination of global and local loss objectives. Specifically, we employ an image-level objective \(L_{\text{DINO}}\) balanced with a patch-level latent reconstruction objective \(L_{\text{iBOT}}\). Additionally, to prevent feature collapse \cite{jing2021understanding} and encourage uniform feature dispersion, a Koleo regularization term is applied. The overall objective for the pretraining stage is given by:
\begin{equation}
L_{\text{pretraining}} = L_{\text{DINO}} + L_{\text{iBOT}} + L_{\text{Koleo}}.
\label{eq:pretraining}
\end{equation}

\noindent\textbf{Gram-Anchoring Post-Training.}  
To enhance the model's ability to encode fine-grained cellular information from HE patches, it is necessary to stabilize and diversify the representations of local tokens. Previous PFMs \cite{chen2024towards,zimmermann2024virchow2}, due to large-scale pretraining, often exhibit inconsistency or collapse in fine-grained local token embeddings rather than progressive improvement during optimization \cite{fan2025scaling}. To address this, we adopt Gram-anchoring \cite{simeoni2025cijo} on a checkpoint obtained from the pretraining stage, introducing an additional term \(L_{\text{Gram}}\) into the loss function. The post-training objective is therefore formulated as:
\begin{equation}
L_{\text{posttraining}} = L_{\text{DINO}} + L_{\text{iBOT}} + L_{\text{Koleo}} + L_{\text{Gram}}.
\label{eq:posttraining}
\end{equation}
This procedure encourages diversity among local token embeddings, resulting in more stable and informative cellular-level representations.

\subsection*{Biomarker Prediction with Cellular-Scale Interpretability} \label{subsec:attention pooling}
\subsubsection*{Problem Definition}
In a vision transformer (ViT)-based \textbf{PFM}, a histopathology image patch $\mathbf{I} \in \mathbb{R}^{H \times W \times 3}$ is divided into a grid of non-overlapping tokens of size $t \times t$. 
These tokens are linearly projected into a sequence of feature embeddings and appended with a learnable class token $\mathbf{z}_{\text{cls}}$ representing global information. The resulting input sequence is defined as:
\begin{equation}
\mathbf{Z}_0 = [\, \mathbf{z}_{\text{cls}}; \mathbf{E}(\mathbf{x}_1), \mathbf{E}(\mathbf{x}_2), \dots, \mathbf{E}(\mathbf{x}_N) \,] + \mathbf{E}_{\text{pos}},
\label{eq:tokenization}
\end{equation}
where $\mathbf{E}(\cdot)$ is the patch embedding function, $\mathbf{E}_{\text{pos}}$ denotes positional encodings, and $N = (H/t)\times(W/t)$ is the number of patch tokens. After $L$ transformer layers, the final hidden representation $\mathbf{Z}_L = [\mathbf{z}^{L}_{\text{cls}}, \mathbf{z}^{L}_{1}, \dots, \mathbf{z}^{L}_{N}]$ is obtained.

\subsubsection*{Attention Pooling}

%
To overcome the limitations of linear probing \ref{linear_probe} and capture hierarchical information across both global and local levels, we introduce an \textbf{attention pooling} mechanism that uses the class token to guide the aggregation of patch-leve embeddings. Specifically, the final representation $\mathbf{h}$ is computed as:

\begin{equation}
\mathbf{h} = \text{Attn}\left(\mathbf{Q}=\mathbf{z}^{L}_{\text{cls}}, \mathbf{K}=\mathbf{Z}^{L}_{\text{patch}}, \mathbf{V}=\mathbf{Z}^{L}_{\text{patch}}\right),
\label{eq:attention_pool}
\end{equation}

where the aggregated feature $\mathbf{h}$ is subsequently fed into a classification head:

\begin{equation}
\hat{y} = \sigma(\mathbf{W}_{\text{attn}} \mathbf{h} + b).
\label{eq:attention_cls}
\end{equation}

This procedure encourages the model to retain cellular-scale morphological details while preserving the global tissue context, leading to more accurate and interpretable biomarker predictions, which will be evaluated in the following section.

\subsubsection*{Visualisation of Fine-grained cellular-scale Heatmap}

To provide cellular-scale interpretability for the model's biomarker prediction,  attention heatmaps are derived from the attention scores $\mathbf{W}_{\text{attn}}$ over all patch tokens $\mathbf{Z}^{L}_{\text{patch}}$. The scores are normalized to $[0,1]$, rearranged according to the spatial patch layout, and upsampled to the original image resolution. The resulting dense importance map is smoothed with a Gaussian filter to reduce block artifacts, thresholded to suppress low-response regions, normalized, and overlaid on the original image for visualization.


\subsection*{Biomarker Prediction Tasks}

To evaluate the clinical robustness and cellular-scale interpretability of our method, we validate our model across ten biomarker prediction tasks, categorized into four functional dimensions: histomorphological biomarkers, molecular biomarkers, genetic alteration biomarkers, and spatial transcriptomic proxy.

\subsubsection*{Histomorphological Biomarkers} \label{morpho_dataset}
To validate Hireca's capacity to identify fine-grained histomorphological biomarkers, we focused on two representative tasks: mitotic figure prediction and tumor budding assessment, which are critical indicators of cellular proliferation and invasive potential.

\noindent\textbf{Mitosis Detection in Multi Organs.}
The recognition of mitotic figures is a 
foundational component of tumor grading and 
serves as a primary prognostic factor for 
oncological outcomes. However, digital mitotic 
count is frequently hindered by significant 
covariate shifts arising from diverse tumor types 
and varying digitization devices. To evaluate 
Hireca's cross-domain robustness, we utilized data 
from the MIDOG 2022 challenge \cite{aubreville2024domain}, 
which encompasses images from multiple scanners 
and tumor organs. 

Following the protocol established by \cite{farooq2024toward}, high-resolution patches centered on expert-annotated mitotic regions were extracted. We formulated mitosis detection as a three-class differentiation task, namely mitotic figures: hard negatives (mitosis-like): negative tissue = 5,550: 8,776: 8,703. Image patches were originally approximately $100 \times 100$ pixels and were resized to $96 \times 96$ pixels before model input. This setup assesses whether Hireca's DINOv3-based features can capture the fine-grained nuclear morphology, to distinguish true mitoses from misleading  cellular debris or apoptotic bodies.

\noindent\textbf{Tumor Budding Detection in Colorectal Cancer.}
In addition to mitosis detection, tumor budding is another important morphology-driven biomarker. This is particularly relevant in colorectal cancer (CRC), where tumor budding, defined as isolated tumor cells or small clusters of fewer than five cells at the invasive front, is closely associated with poor prognosis, metastatic risk, and treatment planning \cite{zhou2026tb}. Therefore, we constructed an internal CRC tumor budding dataset to evaluate whether Hireca can identify these small invasive tumor-cell structures from HE images.

We collected 122 pairs of HE and IHC paired internal CRC WSIs, and they are first spatially aligned using DeepHistReg~\cite{wodzinski2021deephistreg}. Tumor bud candidates are then detected on IHC WSIs using a pre-trained TB-YOLO model~\cite{zhou2026tb}, and the corresponding regions were projected onto the registered HE slides to extract matched HE patches. Since perfect pixel-level alignment between serial sections is challenging, candidate HE patches were further filtered using a pre-trained CRC classification model~\cite{zhao2020artificial}. Finally, expert pathologists reviewed the paired HE-IHC patches and validated the final tumor budding annotations, using the IHC images as reference support for ambiguous HE findings.

The resulting highly curated dataset comprises 5556 paired patches (extracted at $20\times$ magnification with a spatial resolution of $224 \times 224$ pixels), containing 2155 positive tumor budding patches and 3401 negative patches without tumor budding, thereby providing a robust foundation for evaluating biomarker prediction models.


\subsubsection*{Molecular Biomarkers Detection} \label{molecular_dataset}
Compared with histomorphological biomarkers, which are more or less directly visible on HE slides, molecular biomarker prediction is more challenging because the target status is only indirectly reflected in HE morphology. This requires the model to identify subtle morphological correlates of clinically relevant molecular phenotypes. 

\noindent\textbf{Breast Cancer Molecular Biomarkers.}
Breast cancer provides a representative example, with ER, PR, Ki67, and HER2 routinely used for molecular subtyping, treatment selection, and proliferation assessment. We therefore used the publicly available IHC4BC dataset~\cite{akbarnejad2025toward} to evaluate HE-based prediction of these four biomarkers. In our experiments, the dataset contained 20,656 Ki67 patches, 20,713 HER2 patches, 26,134 ER patches, and 24,971 PR patches. Following the dataset protocol, biomarker expression was stratified into four levels from 0 to 3+ using the original expert-validated thresholds. We formulated each biomarker as a binary classification task by grouping 0 and 1+ as negative and 2+ and 3+ as positive. For HER2, the 2+ category was excluded  due to poor inter-observer reproducibility
for 2+ cases.

For direct comparability with the original patch-level benchmark, we followed the dataset split strategy and performed random patch-level splits for each biomarker. Considering class imbalance, patches from each class were split into 80\% training and 20\% testing sets. Splits that contained no instances from any target class were not used. This setup enables evaluation on high-confidence HE-IHC pairs while maintaining consistency with the original dataset protocol.




\noindent \textbf{Pancreas Molecular Biomarkers.}
In addition to breast cancer biomarkers, pancreatic ductal adenocarcinoma (PDAC) represents a clinically important setting for molecular biomarker assessment because of its marked biological heterogeneity and poor clinical outcome. A proprietary internal cohort from the Technical University of Munich (TUM) was therefore used to evaluate Hireca for molecular subtyping of PDAC. 

The dataset, previously characterized in 
\cite{fischer2026contrastive}, consists of spatially aligned HE and IHC 
pairs (HNF1a and KRT81). Following the IHC-based reference standard of the original dataset, biomarker labels were determined by expert pathological examination of the physically stained HNF1a and KRT81 IHC slides, considering both staining intensity and the percentage of stained tumor cells. HNF1a expression was formulated as a binary classification task, separating negative and positive cases. In contrast, KRT81 was treated as a three-class grading task with negative, medium, and strong expression levels, preserving the graded staining patterns observed in the IHC reference. To avoid data leakage, we performed case-level splits, resulting in 168/50 training/testing cases for HNF1a and 167/46 training/testing cases for KRT81, respectively. This setup ensures that 
Hireca learns generalized morphological features that 
correlate with high-dimensional molecular phenotypes 
across different subjects.


\subsubsection*{Genetic Alteration Biomarker}\label{msi_dataset}

Compared with histomorphological and molecular biomarkers, genetic alteration biomarkers are more challenging for HE-based prediction because their histological correlates are often indirect, spatially heterogeneous, and difficult to define by visual inspection. We therefore used microsatellite instability (MSI) prediction as a representative task to evaluate whether Hireca can capture morphology associated with underlying genetic alteration.

\noindent\textbf{Microsatellite instability in gastric cancer.}
MSI is a clinically important biomarker for patient stratification and immunotherapy guidance in gastrointestinal cancer~\cite{kather2019deep,van2019drug}. Although HE-stained whole-slide images are widely available, MSI-associated morphological patterns may be subtle and not consistently recognizable by human experts~\cite{schirris2022deepsmile}. We therefore used a large-scale MSI/MSS classification dataset derived from the TCGA-STAD cohort, as curated in~\cite{kather2019histological}.

This dataset encompasses over 190,000 unique 
FFPE image patches ($224 \times 224$ pixels, 
$0.5 \mu m/px$) categorized into MSS and MSI. 
We maintained the standard patient-level split 
(70\% training, 30\% testing) to prevent 
information leakage. Beyond achieving high 
classification accuracy, the core objective 
of using Hireca here is to leverage its 
fine-grained attention mechanism to visualize 
the morphological hallmarks of genetic 
instability. By generating cellular-scale 
heatmaps, Hireca offers the potential to 
identify the previously ``hidden" histological 
patterns that distinguish MSI from MSS, 
thereby providing biological insights that 
align algorithmic predictions with 
potential cellular-scale defects.

\subsubsection*{Spatial Transcriptomics Proxy biomarker} \label{st_dataset}
Beyond genetic alteration biomarker prediction, spatial transcriptomic state prediction represents an even more challenging setting, as it requires the model to infer spatially resolved gene-expression states that are not directly visible in HE images from local tissue morphology. To establish this evaluation setting, we used our in-house spatial transcriptomics (ST) dataset of mouse pancreas with acinar cell-derived preneoplastic lesions (GEO: GSM8573643, series GSE279507)~\cite{Fischer2025MousePDAC}. The dataset contains FFPE sections from five 6-month-old mice and was profiled using the 10x Genomics Visium platform. RNA profiles were captured within circular spots of $55\mu\text{m}$ in diameter, resulting in a spot-level feature space of 19,465 genes.
Ground-truth transcriptomic niches were defined by combining pathological characterization with unsupervised graph-based clustering in the 10x Genomics Loupe Browser. Specifically, a sparse nearest-neighbor graph and Louvain clustering were used to partition the tissue into $k=8$ clusters~\cite{10xClustering}. These clusters were used as proxy labels for distinct local tissue domains and spatial transcriptomic states.

For each Visium spot, we cropped a corresponding HE image patch centered on the spot coordinate. Patches were extracted from high-resolution HE-stained WSIs at $40\times$ magnification ($\text{MPP}=0.2524$) with a size of $224 \times 224$ pixels. To avoid data leakage, we used a slide-level split: slide 926983 was held out as the independent test set because it contained a representative distribution of the eight transcriptomic clusters, while the remaining four slides, 38032, 926987, 926991, and 926895, were used for training. This resulted in an approximate 80/20 train/test split, with the full patch distribution reported in Supplementary Sec.2.

\subsection*{Comparison Methods}
\subsubsection*{SOTA pathology foundation models}
We exhaustively compare our method with the following several SOTA pathology foundation  methods:

(1) \textbf{ResNet-50} (26M parameters) \cite{he2016deep} is a representative convolutional backbone that uses residual connections to train deep networks effectively. As a widely adopted baseline in medical image analysis, it provides a conventional CNN reference point against newer vision foundation models.

(2) \textbf{DINOv2-L} (307M parameters) \cite{oquab2023dinov2}, published by Meta AI, marked a new milestone in the era of vision foundation models. It was pretrained on 142 million natural images and distilled from a larger teacher model, demonstrating strong generalization across diverse visual tasks.

(3) \textbf{DINOv3-L} (307M parameters) \cite{simeoni2025cijo}, also developed by Meta AI, builds upon DINOv2 by scaling both data and architecture. It was pretrained on 1.7 billion images, introduces more flexible positional embeddings, and employs a post-training regularization strategy to enhance patch-token representations.

(4) \textbf{UNI-L} (307M parameters) \cite{chen2024towards}, proposed by the Mahmood Lab at Harvard Medical School, adapts the DINOv2 self-supervised pretraining framework to digital pathology. It was pretrained on over 100 million patches derived from approximately 100,000 private HE whole-slide images (WSIs), forming one of the first large-scale pathology foundation models.

(5) \textbf{UNI2-H} (681M parameters) extends UNI-L with a larger backbone and expanded data coverage, incorporating over 200 million histopathology patches from more than 350,000 WSIs.

(6) \textbf{Virchow2} (632M parameters) \cite{zimmermann2024virchow2}, introduced by Paige in collaboration with Microsoft, integrates pathology-specific data augmentations, multi-magnification training, and stronger regularization. It was trained on over 3.5 million WSIs using a ViT-H/14 architecture.

(7) \textbf{CONCH1.5} (307M parameters) extends the CONCH framework \cite{lu2024visual} and leverages over 4 million image-caption pairs collected from diverse histopathology datasets and biomedical literature, integrating visual and textual modalities through task-agnostic contrastive pretraining.

(8) \textbf{Phikon-v2} (307M parameters) \cite{filiot2024phikon} is a public DINOv2-based histopathology foundation model trained on 460 million pathology tiles from over 100 public cohorts across more than 30 cancer sites. It was designed as a large feature extractor for biomarker prediction and demonstrated competitive performance among recent pathology foundation models.

\subsubsection*{SOTA Interpretability Methods} \label{linear_probe}
To evaluate the interpretability of our proposed CytoMap, we compare it against two established paradigms of interpretability commonly used for CNN and Transformer architectures:

(1) \textbf{GradCAM} \cite{selvaraju2020grad}: We apply GradCAM to CNN-based baselines (e.g., ResNet-50). GradCAM generalizes the Class Activation Mapping (CAM) \cite{zhou2016learning} by using the gradient information of any target concept flowing into the final convolutional layer to produce a localization map. It highlights the important regions in the image by weighting the feature maps based on their contribution to the model's prediction. As a standard for CNN interpretability, it serves as our baseline for assessing regional-level relevance.
(2) \textbf{Attention Maps} \cite{vaswani2017attention}: For ViT-based foundation models (e.g., UNI-L, Virchow2), the attention weights often capture semantic relationships between patches. Following the standard practice in foundation model evaluation, we extract the attention matrix from the last layer and aggregate the weights across all heads to visualize the model's "focus." This provides a baseline for global/regional attention-based interpretability, which often lacks the cellular-scale granularity that our proposed CytoMap aims to achieve.




\subsubsection*{Cellular-Scale Interpretability Evaluation} \label{interpretability_setup}
\textbf{Heatmap comparison ranks.}
Cellular-scale interpretability is crucial for linking biomarker prediction outcomes  to plausible histopathological evidence. Therefore, we compare the cellular-scale interpretability of our model  with existing methods, including AttentionMap and GradCam. To ensure a fair and clinically relevant assessment, we recruit both four senior pathologists and four junior pathologists from two countries, allowing evaluation across different levels of diagnostic experience while reducing the risk of institution- or training-specific bias. These pathologists independently rank the heatmaps generated by CytoMap(ours), AttentionMap(UNI as backbone), and GradCam (Resnet50 as backbone) in two representative biomarker-related tasks: histomorphological biomarker assessment for mitosis and molecular biomarker assessment for HER2. 

To determine whether the interpretability preference remains stable across experience levels, we quantify the agreement for each task between the senior and junior groups using pairwise agreement. For each case, the group-level rankings of the three models were decomposed into all pairwise comparisons, so that agreement was assessed at the level of relative ordering between every model pair rather than only the top-ranked model. Let $N_i^{\mathrm{agree}}$ denote the number of model pairs for which the senior and junior groups gave the same ordering on case $i$. The image-level pairwise agreement was defined as
\begin{equation}
A_i=\frac{N_i^{\mathrm{agree}}}{\binom{M}{2}},
\end{equation}
where $M$ is the number of models. The overall pairwise agreement was then computed by averaging $A_i$.

\noindent \textbf{Clinical relevance of cellular-scale heatmaps.}
To validate the clinical relevance of cellular-scale heatmaps generated by Hireca, we invited one pathologist from Europe to score each case in HER2 Detection task. The HE image, the predicted heatmap and the corresponding HER2 image are given as reference. The Her2 image was used to verify whether highly activated heatmap regions aligned with biomarker-relevant tumor-cell areas.

Each heatmap was rated on a 1-3 Likert scale across five dimensions: biological plausibility (whether the highlighted regions corresponded to biologically credible tumour-cell or microenvironmental patterns), relevance (whether the highlighted regions were pertinent to HER2 assessment), accuracy (whether biomarker-relevant regions were correctly localized), specificity (whether irrelevant regions were avoided), and clinical usefulness (whether the visualization supported pathological interpretation). Scores of 1, 2, and 3 corresponded to poor, moderate, and good performance, respectively.



\noindent \textbf{Cellular-scale representation analysis of CytoMap-attended tokens.}
We performed a token-level representation analysis to qualitatively assess whether the regions most strongly highlighted by CytoMap corresponded to biologically meaningful cellular and microenvironmental structures. For each task, representative positive and negative patches were selected, and the patch tokens with the highest CytoMap attention scores were identified. Their corresponding Hireca feature embeddings were then extracted for downstream visualization.

Since the effective biological scale differed across tasks, token resolution was adapted accordingly. In the mitosis task, each token covered approximately $32 \times 32$ pixels, corresponding roughly to the scale of a single cell. In the MSI/MSS task, each token covered approximately $16 \times 16$ pixels, reflecting the finer local resolution required in this setting. Token embeddings from positive and negative samples were pooled within each task and projected into two dimensions using t-SNE. Each point in the resulting embedding therefore represented a highly attended local region highlighted by CytoMap. Representative token crops were further mapped back to the embedding space to support pathological interpretation of the observed clusters.

This analysis served as a qualitative representation-level validation of CytoMap. Specifically, it was used to assess whether highly attended regions formed separable structures associated with diagnostic states, thereby supporting the interpretation that CytoMap captures cellular-scale and microenvironment-level diagnostic information rather than only coarse regional patterns.







\subsection*{Role of the funding source}
The funder had no role in study design, data collection, data analysis, data interpretation, writing of the report, or the decision to submit the paper for publication.


\subsection*{Declaration of interests}
The authors have no competing interests to declare that are relevant to the content of this article.

\subsection*{Author contributions}
J.L. and H.L. conceived and initiated the project, designed the study, developed and implemented the methodology, conducted experiments, interpreted the results, and wrote the manuscript. Z.X. contributed to software development, performed experiments, analyzed the results, and assisted in manuscript preparation. F.-L.K., F.St., S.Z., A.K., F.Sc., A.M., J.S., and J.Y. served as expert pathologists, participated in expert evaluations, provided pathological interpretations, and contributed domain-specific pathology insights. T.Z. and X.M. contributed to manuscript revision and figure preparation. C.G., W.Z., M.W., G.J., and H.X. contributed to dataset curation, data management, and manuscript writing. B.L. contributed to data analysis and provided scientific feedback. S.K.Z., N.N., and P.S. supervised the project, provided scientific guidance, and substantially revised the manuscript. P.S. additionally contributed to manuscript writing and project coordination. All authors reviewed and approved the final manuscript.

\subsection*{Acknowledgements}
This work was partly supported by the BMFTR-funded SATURN3 project
(01KD2206C) and the IMI BIGPICTURE project (IMI945358).


\bibliographystyle{vancouver}
\bibliography{sn-bibliography}

\newpage{}

\end{document}